\documentclass{article}

\pdfoutput=1
\usepackage{arxiv}

\usepackage[utf8]{inputenc} 
\usepackage[T1]{fontenc}    
\usepackage{hyperref}       
\usepackage{url}            
\usepackage{booktabs}       
\usepackage{amsfonts}       
\usepackage{nicefrac}       
\usepackage{microtype}      
\usepackage{lipsum}		
\usepackage{graphicx}
\usepackage{doi}
\usepackage{amsmath}
\usepackage{placeins}
\usepackage[square,sort,comma,numbers]{natbib}
\usepackage{algorithm}
\usepackage{algpseudocode}

\title{Unsupervised Spiking Instance Segmentation on Event Data using STDP}

\author{ Paul~Kirkland,~Davide~L.~Manna,~Alex~Vicente-Sola,
        and~Gaetano~Di~Caterina
        \\
	Neuromorphic Sensor Signal Processing Lab\\ Centre for Image and Signal Processing\\ Electrical and Electronic Engineering\\ University of Strathclyde\\ Glasgow, United Kingdom.\\
	\texttt{paul.kirkland@strath.ac.uk} \\
}

\renewcommand{\shorttitle}{USIS-STDP}

\hypersetup{
pdftitle={Unsupervised Spiking Instance Segmentationon Event Data using STDP},
pdfsubject={q-bio.NC, q-bio.QM},
pdfauthor={Paul Kirkland},
pdfkeywords={Neuromorphic Engineering},
}

\begin{document}
\maketitle

\begin{abstract}
	Spiking Neural Networks (SNN) and the field of Neuromorphic Engineering has brought about a paradigm shift in how to approach Machine Learning (ML) and Computer Vision (CV) problem. This paradigm shift comes from the adaption of event-based sensing and processing. An event-based vision sensor allows for sparse and asynchronous events to be produced that are dynamically related to the scene. Allowing not only the spatial information but a high-fidelity of temporal information to be captured. Meanwhile avoiding the extra overhead and redundancy of conventional high frame rate approaches. However, with this change in paradigm, many techniques from traditional CV and ML are not applicable to these event-based spatial-temporal visual streams. As such a limited number of recognition, detection and segmentation approaches exist. In this paper, we present a novel approach that can perform instance segmentation using just the weights of a Spike Time Dependent Plasticity trained Spiking Convolutional Neural Network that was trained for object recognition. This exploits the spatial and temporal aspects of the network's internal feature representations adding this new discriminative capability. We highlight the new capability by successfully transforming a single class unsupervised network for face detection into a multi-person face recognition and instance segmentation network.
\end{abstract}


\section{Introduction}
Human vision has the innate ability to detect, localise, and differentiate objects of interest, even in the face of multiple background and foreground distractors. Meanwhile, in Computer Vision (CV), this ability known as instance segmentation is a complex and challenging task. Instance segmentation is the detecting and pixel-wise delineation of each distinct object of interest appearing in an image. In essence instance segmentation is the hybrid of two key computer vision tasks, object detection and semantic segmentation. Object detection is the process of detecting instances of objects belonging to a certain class, while also identifying their spatial location typically with a bounding box. While semantic segmentation is the task of clustering parts of an image together that belong to the same object class, resulting in a much more detailed pixel-wise localisation. 

The main component of an object detector for segmentation is a good internal feature representation \cite{girshick2014rich,dickinson2009object}. Traditionally, considerable effort was put into designing handcrafted local descriptors: SIFT \cite{lowe1999sift,lowe2004siftd}, HOG \cite{dalal2005hog} and SURF \cite{bay2006surf}. These low-level features are then utilised to create high-level image representations through Bag of Words \cite{sivic2003bowvideo} and Fisher Vectors \cite{perronnin2010improvingfisher}. Which in turn, are then used for classification. The downside to these feature representation methods is that often the internal feature representation was based on the level of domain expertise. Lately, with the advent of Deep Learning (DL) \cite{Lecun2015deep}, beyond state-of-the-art performance \cite{Voulodimos2018deep} was now reachable without any domain expertise and handcrafted features for a multitude of tasks: image recognition (AlexNet and ResNet) \cite{Krizhevsky2012alexnet, He2016resnet},  object detection (Faster R-CNN and RFCN) \cite{Ren2015fasterrcnn, Dai2016rfcn}, object tracking (FCNT and MD Net)\cite{Wang2015fcnt,Nam2016MDnet} and semantic and instance segmentation (FCN, SegNet, U-Net and Mask R-CNN) \cite{Long2015,Badrinarayanan2017,Ronneberger2015unet,He2017mask}. This improvement is mainly due to the combination of feature extraction and classification now being governed by supervised techniques using backpropagation. Consequently, this shifted the problem of internal feature representation to the development of bigger and better neural networks, where the goal of better segmentation accuracy often transcends other factors, such as segmentation efficiency. This leads to a stark increase in the real-time computational and memory overhead of the system \cite{Marcus2018deep, Thompson2020computational}. Not to mention the burden of large volumes of labelled pixel-wise segmentation images for training. This is especially true for deployable systems, where the tight constraints on computation are most prevalent \cite{Garcia2019estimation}, while also requiring an accurate, real-time application. This often leads to an impasse in terms of efficiency and accuracy, with the traditional methods being more efficiency focused, while the DL methods being more accuracy focused. Coupled with these, other desirable features of deployable systems are adaptability and continual learning \cite{zenke2017continual}, allowing the system to better deal with changes in the environment and unprecedented situations.

The expectation of a modern deployable system is threefold \cite{chen2019edgereview}: low latency, in its ability to perform accurate real-time calculations; low power, in its efficiency in implementing the computations; continual learning, in having the ability to learn new features or objects on the fly. Given this, one might expect a complex method is required to achieve all these objectives in one system. However, Neuromorphic Engineering \cite{indiveri2011frontiersNM,mead2020NM}, exploiting its paradigm shift from utilising Spiking Neural Networks (SNN) \cite{ghosh2009spiking} and Event-Based Vision Sensors (EBVS) \cite{Gallego2020evbsurvey}, has demonstrated that such a system can be attainable. Furthermore, we demonstrate that from a simple unsupervised Spike Time Dependent Plasticity (STDP) \cite{Bi1998} trained objection recognition Spiking Convolutional Neural Network (SCNN) \cite{Kheradpisheh2018,masquelier2007unsupervised}, our method can extend this to object detection and instance segmentation. Additionally, this occurs without any further learning, so in essence, can be strapped onto any STDP trained network.

Our proposed method, called Hierarchical Unravelling of Linked Kernels and Similarity Matching through Active Spike Hashing (HULK SMASH), presents a technique of utilising the internal featural-temporal representation to gain this instance segmentation ability, that is using the changing internal feature representation over time to further discriminate (i.e. sub-cluster) the intra-class representations. HULK SMASH utilises the inherent saliency of the STDP spatial-temporal features captured by the original network for each layer. Meaning it no longer just classifies from the upmost hierarchical layer of the SCNN. Instead utilising the temporal occurrence of internal feature representations to take an unsupervised network trained to recognise one class, faces, and extend it to identify and localise individuals. This idea builds on the efficient use of low-level features of traditional CV methods, while using networks and techniques from DL to build better hierarchical features and achieve better accuracy. SNNs provide the ideal platform for this hybrid approach, while further extending the feature set of the implemented network.

The HULK SMASH algorithm is not only able to exploit all the useful features of Neuromorphic Engineering, but adds a further level of interpretability. This is in part due to the use of sparsity of the visual features making them easy to realise, while also exploiting the saliency mapping of the SCNN to help visualise which pixels and features played a part in the classification process. This allows the overall interpretability of the network to increase, with the decisions made by the network to become viewable and assessable, so removing the black box stigma of neural networks. 
This work makes use of previous research  that showed how an SCNN trained with STDP could first perform image classification \cite{Kirkland2019uav}, and recently how it can be transformed into an encoder-decoder network, allowing semantic segmentation to be realised \cite{Kirkland2020spikeseg,kirkland2020pua}. However, this previous research could not distinguish instances within these semantic segmentations. Therefore, the main contributions of this paper are:
\begin{itemize}
    \item The instance-wise mapping of semantic labels through the decoder network, HULK. This allows the individual spiking instances of the deepest convolutional layer (the pseudo classification layer) of a fully convolution encoder-decoder to be tracked back to the pixel domain. This tracking illustrates that each semantic label is often built up from multiple spiking instances from within the pseudo classification layer. 
    \item These semantic instances are then passed onto SMASH, which looks at the network’s internal featural-temporal representation and uses this to compare instances. Allowing a metric in which to form an evaluation of similarity between instances. This is then coupled with a locality checking algorithm which then reports a SMASH score for each instance. Where similar instances can combine to form larger representations for objects. Failing that, creating a new object for any dissimilar instances.
\end{itemize}

\noindent The formulation of the algorithms HULK and SMASH is described in Section 2, illustrating how to extrapolate the internal feature representation from a network. Section 3 details the experimental results used to gauge the ability of the network to now perform object detection and instance segmentation. This is also extended into situations where partial and full occlusions of objects happen, with the final experiment showing how the network performs on occlusion recovery. Section 4 presents details on the interpretability of the network. Discussing how visualising the internal feature representation of the network can help to explain how the original object recognition network was extended for object detection and instance segmentation task. Section 5 then concludes the paper.

\section{Temporal Spike Matching}
\label{ch:TSM:sec:overview}
\begin{figure}[hbt!]
\begin{center}
\includegraphics[width=0.7\linewidth]{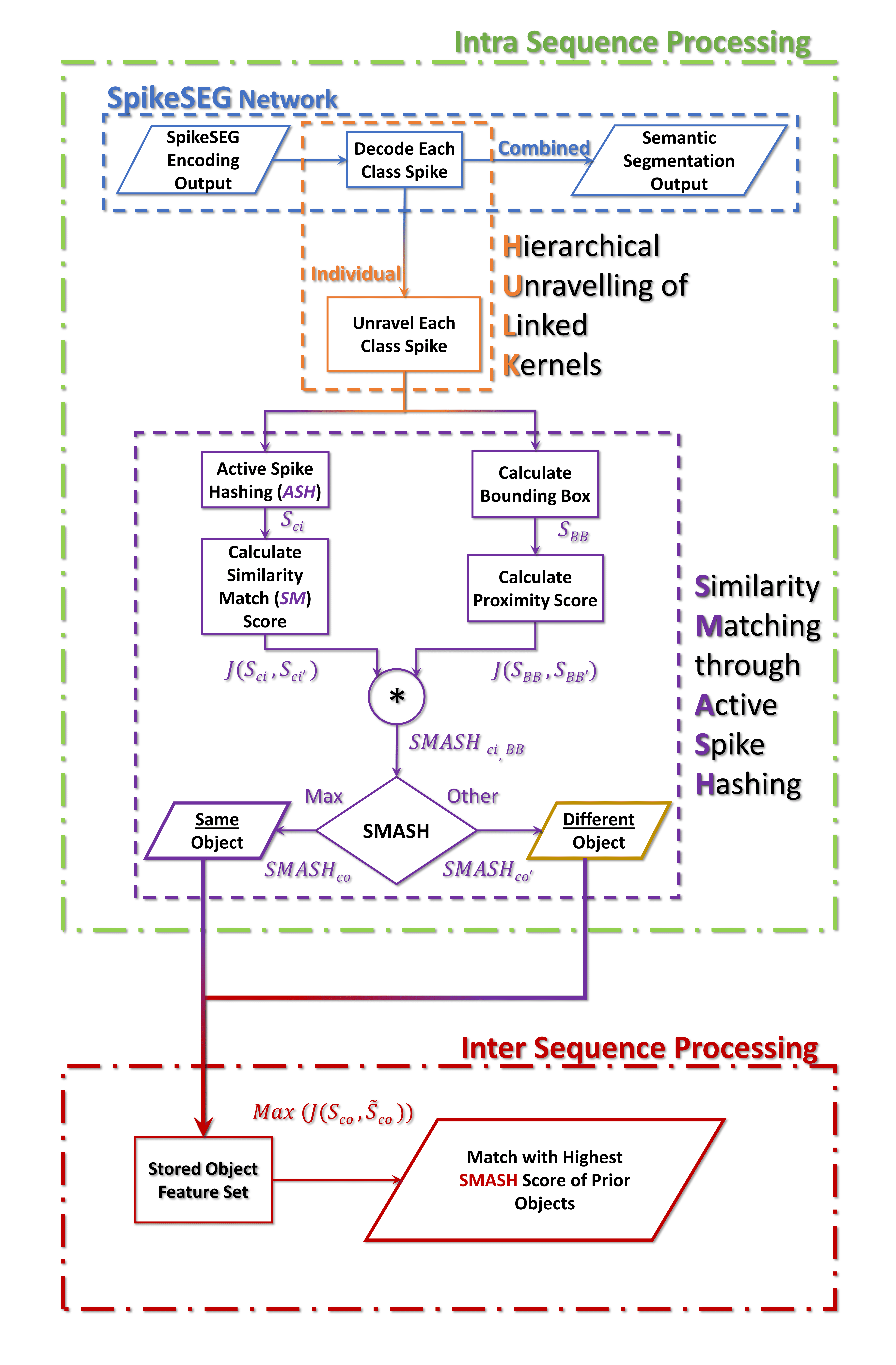}
\end{center}
\caption{Block Diagram of the HULK SMASH system and where it intersects with the SpikeSEG Network} 
\label{fig:SMASHflow}
\end{figure}

The proposed temporal spike matching algorithm HULK SMASH makes use of an image classification SNN trained using STDP, similar to that seen in \cite{masquelier2007unsupervised,Kheradpisheh2018,Kirkland2019uav}. Prior developments have shown that extending an image classification SNN to a semantic segmentation network was possible \cite{kirkland2020pua,Kirkland2020spikeseg}. These two prior works also utilised the N-Caltech dataset \cite{Orchard2015converting} making use of the event driven nature of the input sequences. This work presents how semantic segmentation can also be further extended into an efficient method for object detection and instance segmentation, 
where instead of using a regression process to indicate which classification instance are connected, as seen with DL models \cite{Long2015,Badrinarayanan2017,Ronneberger2015unet,He2017mask}, a proximity and temporal-featural (time- and feature-based) similarity metric is used. This allows the temporal occurrence of the internal feature representation, brought about by the temporal dynamic of the input event stream, to further discriminate intra-class instances.
As illustrated in \figurename{} \ref{fig:SMASHflow}, this method of temporal spike matching for instance segmentation can be broken down into two main parts, the \emph{Intra-} and \emph{Inter-Sequence Processing}: in other words, what is happening internally and externally throughout the process. The intra-sequence is where most of the processing happens, while the inter-sequence allows the processing to link to other instance of the process running, e.g. to compare object instance to see if they are the same object.
This can be seen in a block diagram of the proposed method in \figurename{} \ref{fig:SMASHflow}, within the green and red dot-dashed boxes. 

\figurename{} \ref{fig:SMASHflow} highlights that the \emph{intra-sequence Process} starts at the intersection of the encoder decoder network from SpikeSEG \cite{Kirkland2020spikeseg}. The first step of the process is to individually process each of the spiking instances from the classification layer of the SpikeSEG network. This is in contrast to the original network, which just grouped all instance based on class. The process of decoding each instance individually is what is referred to as the HULK process. Once each classification spike has been decoded back into the original pixel space, this can then be feed to the SMASH process. This in turn calculates a bounding box of each instance $S_{BB}$ in the pixel space, by taking the max and min value in the x and y coordinates. This bounding box is then it used as a \emph{Proximity Score} check with every other instance's bounding box $S_{BB'}$ within the \emph{intra-sequence}. This \emph{Proximity Score} is calculated with the Jaccard Index  $J(S_{BB}, S_{BB'})$ \cite{Jaccard1901distribution}.
Meanwhile, the other parallel process performs the \emph{Active Spike Hashing (ASH)} part of the SMASH process, which stores 2D featural-temporal data $S_{ci}$ from the 4D spatial-featural-temporal decoding class instance processing (X, Y, Feature Map and Time). This then passes onto the \emph{Similarity Matching (SM)} phase, where a similarity score is given to the featural-temporal data compared with every other instance $S_{ci'}$ within the \emph{intra-sequence}, again with the Jaccard index $J(S_{ci}, S_{ci'})$. 
The combination of both the Similarity and Proximity Score gives the SMASH score $SMASH_{ci,BB}$. This indicates if the $S_{ci}$ and $S_{BB}$ of that particular class instance match any other $S_{ci'}$ and $S_{BB'}$, resulting in an outcome of whether the instance is part of the same \emph{Object} $SMASH_{co}$ or a different one $SMASH_{co'}$. This intra-sequence process is run in parallel once all the instances have their ASH and bounding box process complete. This is run for every input sequence of data. Which in this case is 10ms of NVS N-Caltech101 \cite{Orchard2015converting} event data, which is internally treated as an asynchronous stream based on the timestamps of the events.
This allows both object detection and instance segmentation to be performed on this spiking input sequence.

\subsection{Hierarchical Unravelling of Linked Kernels}
\label{ch:TSM:sec:HULK}
\begin{figure}[hbt!]
\begin{center}
\includegraphics[width=0.85\linewidth]{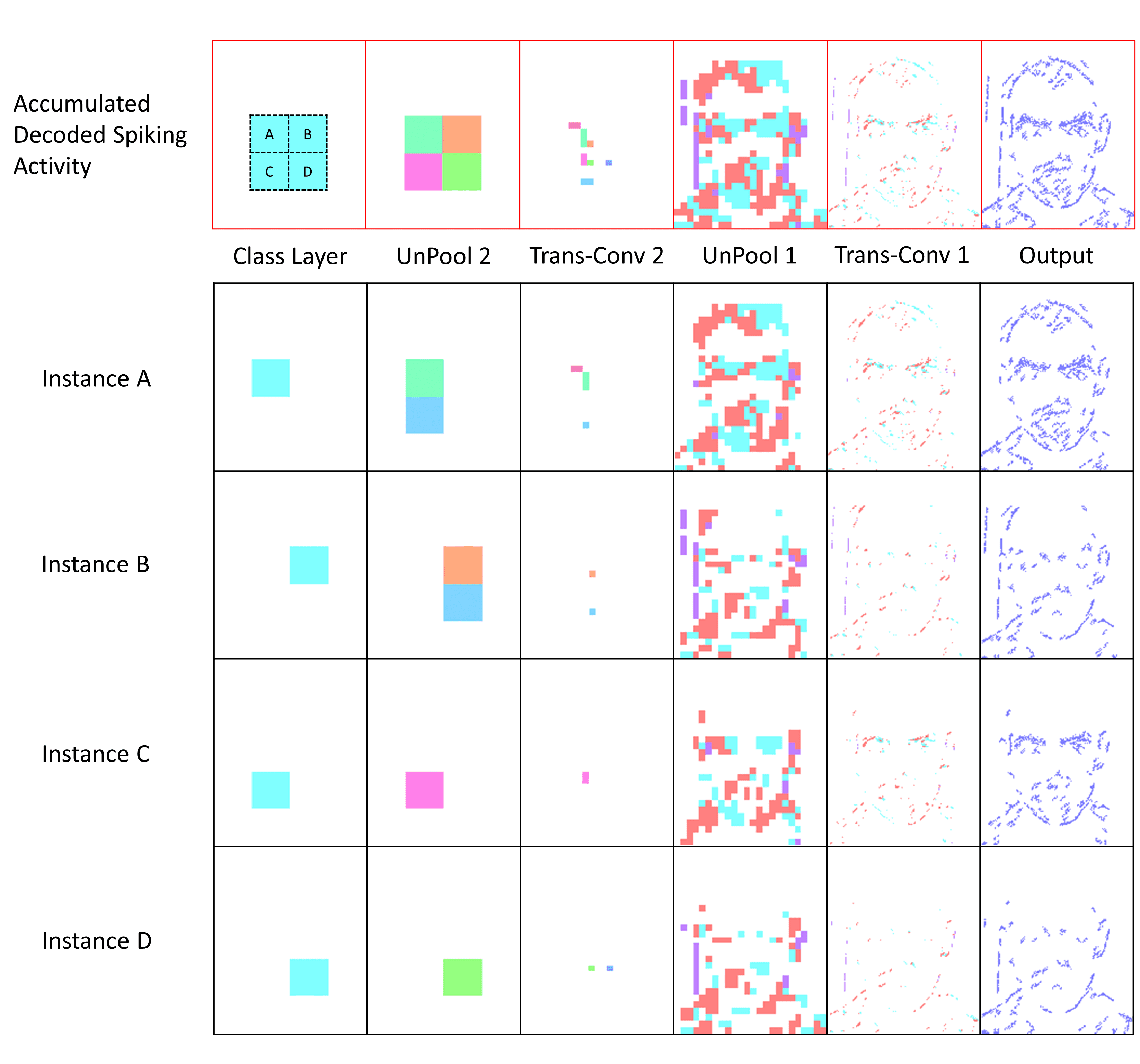}
\end{center}
\caption{Illustration of the individual class spikes being unravelled, indicating activity caused by these class spikes} 
\label{fig:HULK}
\end{figure}

The Hierarchical Unravelling of Linked Kernels (HULK) process permits spiking activity from the classification convolution layer, to be tracked as it propagates through the decoding layers of the network. This tracked propagation allows a record of each subsequent layer's spatial (x, y), featural (m) and temporal (t) spiking activity. 
This process runs in parallel with the transposed convolution process, which itself is mapping each active spike in a layer to the subsequent layer. However, after each layer transform, a record of each spike
is stored. This allows a causal hierarchical map to be created, tracing each classification spike back to the original pixel space.
This process is illustrated within \figurename{} \ref{fig:HULK}, with a decoded sequence shown entitled `Accumulated Decoded Spiking Activity' and the individual class spike breakdown from this are shown within Instance A through D. Each instance, in this case, is representing a single spiking pixel from the classification layer. Whereas in semantic segmentation, all the instance belonging to one class are treated as the same entity and decoded. \figurename{} \ref{fig:HULK} highlights how the class instance is broken down into its individual instances, where each instance often provides enough information to recreate the face in the output pixel space. Although, it is clear that some of the instances favour certain features over others. It is also apparent that there is much repetition and accumulation of features even with this sparse spiking domain.
It is through this process of unravelling the classification spiking activity that permits the ensuing similarity matching process.

\FloatBarrier

\subsection{Similarity Matching through Active Spike Hashing}
\label{ch:TSM:sec:SMASH}
The Active Spike Hashing (ASH) is the process of taking the recorded spiking activity and implementing an efficient and effective way to store the sparse 4 dimensional spatial (x, y), featural (m) and temporal (t) values. 
This is done by realising that the convolution structure of the SCNN SpikeSEG is already dealing with the translational invariance and spatial dimension, together with the bounding box proximity score. 
It can also be noted that position within the scene is not a useful evaluation metric of the similarity between two objects.
The ASH process then results in a 2D featural-temporal hashing of the spiking activity. The featural data then has the same total number of features as the network, which in this case, is 41. This value comes from the use of a network with a similar structure as the Face-Motorbike Network used Kirkland et al. \cite{kirkland2020pua}, except with only one classification layer. The other 40 features come from the 36 from Trans-Conv2 and 4 from Trans-Conv1.

A further memory reduction to the ASH process is permitted by storing the values as binary terms. As the number of spikes recorded within each map per timestep is more a measure of the total spiking activity, rather than the featural-temporal characteristics we evaluate with the similarity measure.

\begin{figure}[hbt!]
\begin{center}
\includegraphics[width=0.85\linewidth]{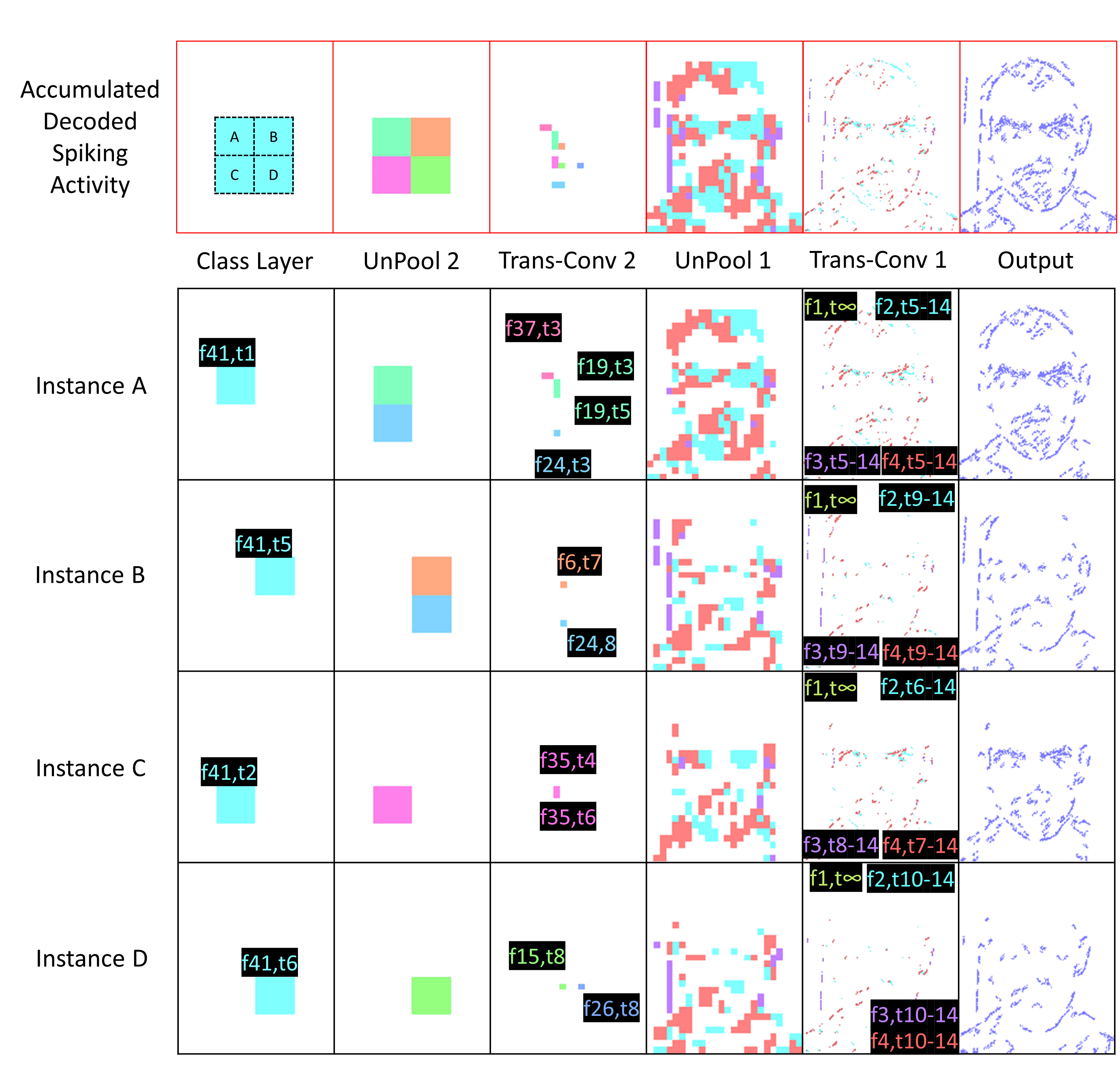}
\end{center}
\caption{Further explanation of the Active Spike Hashing mechanism from the HULK process, feature map and time given to each active neurons} 
\label{fig:HULK_ASH}
\end{figure}

An illustration of the ASH process where the spiking activity is assigned its feature map and time-stamp index (f,t) can be seen in \figurename{} \ref{fig:HULK_ASH}. This builds upon the HULK process previously seen in \figurename{} \ref{fig:HULK} with the indices of the spiking activity now visible in \figurename{} \ref{fig:HULK_ASH} for each convolutional layer. 
Each active neuron is being assigned an f and t index on a class-instance basis. Allowing a 2D matrix of each instance to be realised where from \figurename{} \ref{fig:HULK_ASH}, Instance A would have 1s in the first column (time) forth row (feature map) for the classification layer activity. Meanwhile, the green coloured features of Instance A's Trans-Conv 2, are stored with row 19 and column 3 and 5. The ASH process typically reduces the size of the tensor by 98\%, reducing the memory overhead considerably. The result of the ASH process leaves the internal feature representation similar to that of a conventional spike train.  

Once the spiking activity is hashed, it is ready for the Similarity Matching section of SMASH. The Jaccard similarity coefficient was once again utilised, within the intersection over union used to calculate the segmentation bounding box overlap. However, within this instance, the measure is used to see how much similarity there is from each instance in the intra-sequence process, comparing the featural-temporal similarity of each instances spiking activity, with respect to the overall amount of spiking activity \cite{Jaccard1901distribution}. 

\begin{equation}
\label{EQ:Jac-Sim}
J(S_{Ci},S_{Ci^{\prime}}) = \frac{|S_{Ci}\cap S_{Ci^{\prime}}|} {|S_{Ci} \cup S_{Ci^{\prime}}|}
\end{equation}

\noindent where $S_{Ci}$ is the spiking activity of that class instance, and $S_{Ci^{\prime}}$ represents the spiking activity of any other class instance for that sequence.
However, due to the ASH process storing binary values of the activities, this equation can be simplified down to a logical calculation performed with only ORs and ANDs. This allows the quick comparison of the number of spikes that feature in both the current instance \emph{and} the comparison, divided by the number of spikes that feature in the current instance \emph{or} the comparison. 

\begin{equation}
\label{EQ:Bin-Sim}
J(S_{Ci},S_{Ci^{\prime}}) = \frac{S_{Ci} \land S_{Ci^{\prime}}} {S_{Ci} \lor S_{Ci^{\prime}}}
\end{equation}

\noindent To complete the intra-sequence of the SMASH process, the bounding box IoU score must be calculated for each class instance bounding box $Ci_{BB}$ against the other instances bounding boxes $Ci_{BB^{\prime}}$, instead of for each class in total

\begin{equation}
\label{EQ:classJac-BB}
J(Ci_{BB},Ci_{BB^{\prime}}) = \frac{|Ci_{BB}\cap Ci_{BB^{\prime}}|} {|Ci_{BB} \cup Ci_{BB^{\prime}}|}
\end{equation}

\noindent Multiplication of the similarity score, with the IoU, results in the novel proposed SMASH score for each instance

\begin{equation}
\label{EQ:Smash}
\text{SMASH}(Ci, Ci^{\prime}) = J(S_{Ci},S_{Ci^{\prime}}) \times J(Ci_{BB},Ci_{BB^{\prime}})
\end{equation}

\noindent Once the SMASH score is calculated for each instance, the maximum of each instance is assigned to a class object $Co$ 
\begin{equation}
\label{EQ:Object}
S_{Co} = \operatorname*{arg\,max}_{Ci} (\text{SMASH}(Ci, Ci^{\prime}))
\end{equation}

\noindent This maximum SMASH score is based on the pairing $Ci, Ci^{\prime}$, where each object $S_{Co}$ is compiled from any overlapping parings e.g. for 5 instances the argmax presents the following sets of $Ci, Ci^{\prime}$ pairs, (1-3), (2-4), (3-5), (4-2), (5-3). The values 2 and 4 are assigned to one object, while 1, 3 and 5, as 3 and 5 are matched pairs and 1 is associated with 3, so becomes part of that object.
It is these class objects $S_{Co}$, that are used within the inter-sequence processing. This permits class objects from preceding sequences $\tilde{S_{Co}}$ to be compared with current ones, again looking for similarity with the binary Jaccard coefficient 

\begin{equation}
\label{EQ:Object-Sim}
J(S_{Co},\tilde{S_{Co}}) =\operatorname*{arg\,max}_{S_{Co}}  \frac{S_{Co} \land \tilde{S_{Co}}} {S_{Co} \lor \tilde{S_{Co}}}
\end{equation}

\noindent These class objects within the inter-sequence then allow sequence to sequence continuity, thus allowing tracking of objects permitted that they maintain a level of feature similarity.

As the original SpikeSEG network outputs a semantic segmentation output, the HULK and SMASH processes are supplementary to this process in transforming the semantic regions into instance-based objects. Based on the block diagram seen in \figurename{} \ref{fig:SMASHflow}, the pseudo-code for the SMASH process is provided within Algorithm \ref{CHalgorithm}. This pseudo-code allows further insight to the method for comparison both intra- and inter-sequence and serves to compliment the block diagram seen in \figurename{} \ref{fig:SMASHflow}, but with the addition on the internal functions used to calculate the values.

\begin{algorithm}
\caption{SMASH}
\label{CHalgorithm}
\begin{algorithmic}[1]
\Procedure{intra-sequence}{$a,b$}
\For {\textit{ each input sequence}}
	\For{\textit{each spiking instance in the classification layer}}
		\State \textit{perform HULK for each instance}
		\State \textit{compute max and min, x and y values}
		\State \textit{compute the [$X, Y, W, H$] bounding boxes}
		\State \textit{perform ASH to create $S_{Ci}$}
		\State \Return{$S_{Ci}$}
	\EndFor
	
	\For{\textit{each hashed instance, $S_{Ci}$}}
		\State \textit{compute Similarity score against other instances, $S_{Ci^{\prime}}$  as in 
		(\ref{EQ:Bin-Sim}) }
		\State \textit{compute Bounding Box IoU score against other instances $S_{Ci^{\prime}}$ as in (\ref{EQ:classJac-BB})}
			\State \textit{compute the SMASH score as in (\ref{EQ:Smash})}	
			\State \textit{assign max SMASH score instance pair to object $S_{Co}$ as in (\ref{EQ:Object})}
			\State \Return{$List of S_{Co}$}
	\EndFor
\EndFor

\EndProcedure
\Procedure{inter-sequence}{$a,b$}
	\For{each class object $S_{Co}$}
		\State \textit{compute Similarity score against previous class objects $\tilde{S_{Co}}$ as in (\ref{EQ:Object-Sim}) }
		\State \Return $max$($J(S_{Co},\tilde{S_{Co}})$)
	\EndFor

\EndProcedure
\end{algorithmic}
\end{algorithm}

From the pseudo-code, it can be determined that the SMASH method has two conditions before it concludes that instances are of the same object:
\begin{itemize}
	\item That the two instances must have overlapping bounding boxes
	\item That the two instances must share a featural-temporal data
\end{itemize}
Without either of these, the multiplication of the similarity score and intersection over union will result in 0, and it will determine that the two instances are different objects. All instances that combine into the same object are then stored as one set of features and used at the inter-sequence stage to track objects over multiple input sequences.

The HULK and SMASH processes are able to build upon the semantic segmentation abilities of the semantic segmentation network. Giving object detection and instance segmentation capability, without the use of a regression sequence. Instead, this method opts for locality and similarity of the most salient features, utilising the information already available from the SCNN process.


\section{Experimental Results}
\label{ch:TSM:sec:results}
In the evaluation of the performance of the novel HULK SMASH algorithm presented in this paper, a number of tests have been carried out. The tests include:

\begin{itemize}
	\item Semantic to instance segmentation with object detection
	\item Sequence to sequence tracking of objects with object occlusion and recovery
	\item Similarity matching for intra-class grouping (feature detection)
\end{itemize}

\noindent These tests are all carried out with the N-Caltech Dataset \cite{Orchard2015converting}, mostly utilising the Face category. This Face subset provides a subtle yet distinct amount of intra-class variance, allowing the successful extraction of multiple diverse face life features within the second convolution layer, which then provides a robust and general face detection within the classification layer. The training was carried out using 10 of the different people within the subset, with testing being carried out on the same 10 people, however with different input sequences used for each. A total of 3 input sequences are used in training, each containing 300ms of spiking activity, so 30 buffered inputs (3 of each of the 10 people), for a total of 900 input sequences. Testing was carried out using 2 input sequences, so 600 buffered inputs. 
The network parameters are set to be the same as within \cite{Kirkland2020spikeseg}, with the only difference being 36 features available for the one class present. This was to limit the number of external factors and focus the testing on the intra-classification abilities rather than inter classification. To further validate the SMASH method the 5 and 10 class networks where also utilised from \cite{kirkland2020pua}. These tests helps to validate the SMASH method in a more complex feature similarity environment, where each class has less representation in layer Conv 2, as these networks utilised 16 features times the number of classes, 80 and 160 for the 5 and 10 classes respectively. 

\subsection{Semantic to Instance Segmentation with Object Detection}
\begin{figure}[hbt!]
\begin{center}
\includegraphics[width=0.98\linewidth]{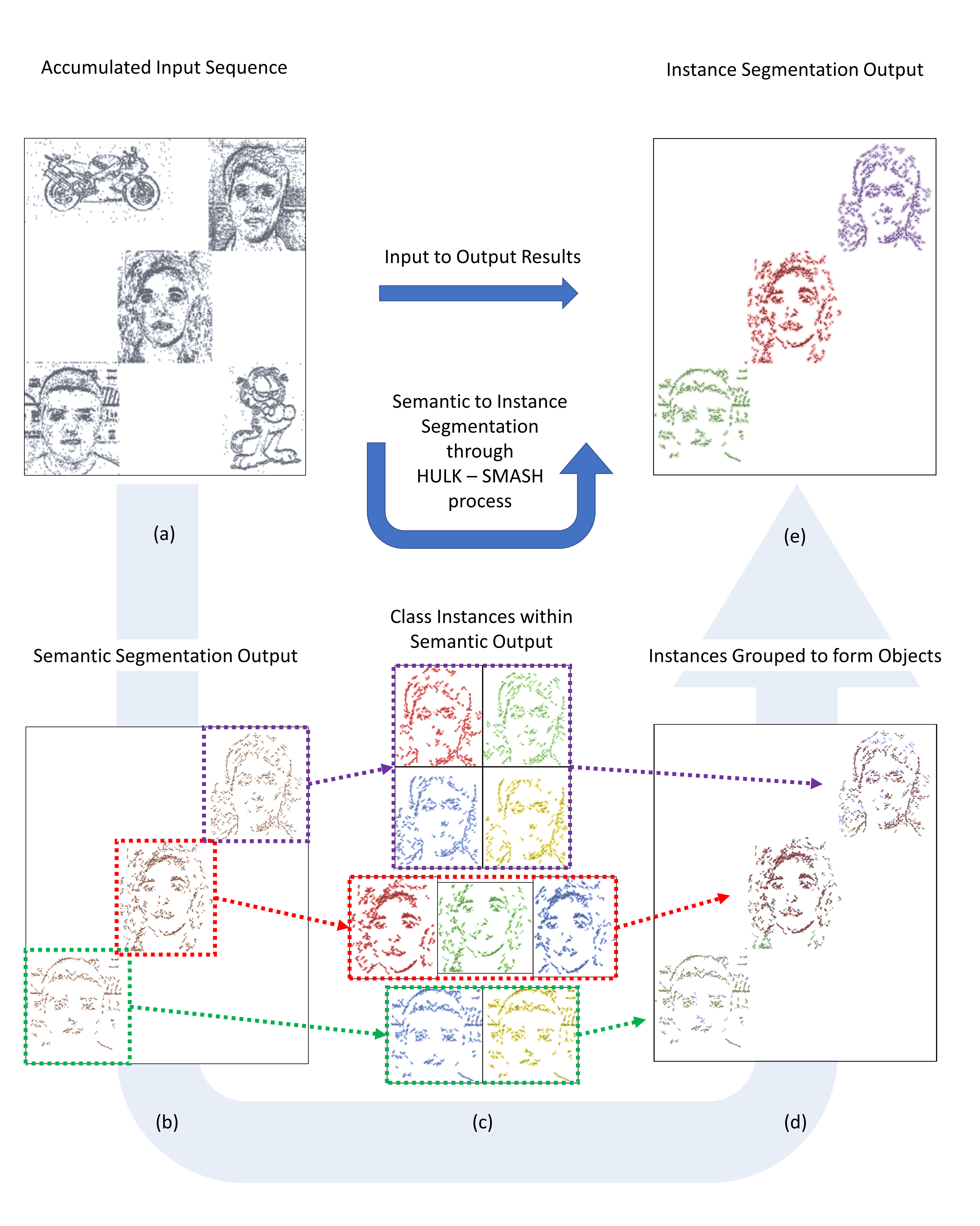}
\end{center}
\caption{Processing from Semantic Segmentation to Instance Segmentation: (a) the accumulated input sequence, (b) the semantic segmentation output, (c) the class instances from the semantic segmentation, (d) the grouped class objects and (e) the instance segmentation output } 
\label{fig:instanceseg}
\end{figure}

A series of multiple input streams are given as input to the SpikeSEG network, illustrated in \figurename{} \ref{fig:instanceseg}. The network then creates both the semantic segmentation output and via the HULK-SMASH process, an object detection/instance segmentation output.
The input sequence is shown in \figurename{} \ref{fig:instanceseg} (a), with 5 input stream presented at the same times, 3 with Faces and 2 without. \figurename{} \ref{fig:instanceseg} (b) highlights the semantic segmentation output of the SpikeSEG network, while \figurename{} \ref{fig:instanceseg} (c) demonstrates the HULK process extracting each instance from the semantic representation. \figurename{} \ref{fig:instanceseg} (d) illustrates how the SMASH process groups instances that score above 0 into class instance objects. Lastly \figurename{} \ref{fig:instanceseg} (e) presents the final instance segmentation output, thanks to the object detection separating the semantic representation.

Within this scenario shown in \figurename{} \ref{fig:instanceseg}, the separate input streams are not overlapping, consequently meaning the bounding boxes between input stream also do not overlap. This means the SMASH score between the input streams will always be 0 regardless of the similarity score. This is partially the intention of the SMASH process, allowing objects that do not share spatial locality to lead to false object grouping. As such the performance of the object detection to identify that the individual class instances belonging to one object is being evaluated, similar to the process depicted in \figurename{} \ref{fig:HULK} and \ref{fig:HULK_ASH}. This means that even though the network is classifying all the Faces as just that, Faces, the SMASH process is able to identify that these do no share the same featural content. The object detection accuracy to identify three separate distinct faces is 100\% over the complete input sequence, with numerical results tabulated within Table \ref{tab:my-table}. However, over each input buffer, this drops to 97.33\% as there are a few cases in which the similarity is 0 due to a low input spiking rate producing the minimal number of spikes for activation. This is typical of the inputs from the start and end of the saccade movement within the N-Caltech dataset \cite{Orchard2015converting}.

\begin{table}[]
\caption{Results from the experimental testing split into 4 sections covering the Object detection, detection with occlusion, occlusion recovery and object self matching. *MS - Multistream}
\label{tab:my-table}
\resizebox{\linewidth}{!}{%
\begin{tabular}{cccl}
\textbf{}                                 & \multicolumn{3}{c}{\textbf{Object Detection Accuracy (\%)}}                                         \\  
\textbf{}                                 & \textbf{per Input}            & \textbf{per Sequence}         &                                     \\ \hline
\textbf{MS N-CalTech (Face)}     & 97.33                         & 100                           &                                     \\
\textbf{MS N-CalTech (5 Class)}  & 95.17                         & 100                           &                                     \\
\textbf{MS N-CalTech (10 Class)} & 91.67                         & 100                           &                                     \\ \hline
\textbf{}                                 &                               &                               &                                     \\
\textbf{}                                 & \multicolumn{3}{c}{\textbf{Detection Accuracy with rate of Occlusion (\%)}}                         \\  
\multicolumn{1}{l}{}                      & \textbf{5\%}                  & \textbf{25\%}                 & \multicolumn{1}{c}{\textbf{50\%}}   \\ \hline
\textbf{MS N-CalTech (Face)}     & 97.33                         & 62                            & \multicolumn{1}{c}{28}              \\ \hline
\textbf{}                                 & \multicolumn{1}{l}{\textbf{}} & \multicolumn{1}{l}{\textbf{}} & \textbf{}                           \\
\textbf{}                                 & \multicolumn{3}{c}{\textbf{Occlusion Recovery Rate (\%)}}                                           \\
\multicolumn{1}{l}{}                      & \textbf{without noise}        & \textbf{with noise}           & \multicolumn{1}{c}{\textbf{}}       \\ \hline
\textbf{MS N-CalTech (Face)}     & 100                           & 100                           & \multicolumn{1}{c}{}                \\ \hline
\multicolumn{1}{l}{}                      & \multicolumn{1}{l}{\textbf{}} & \multicolumn{1}{l}{\textbf{}} & \textbf{}                           \\
\textbf{}                                 & \multicolumn{3}{c}{\textbf{Object Self Matching}}                                                   \\ 
\multicolumn{1}{l}{}                      & \textbf{Top 1}                & \textbf{Top 3}                & \multicolumn{1}{c}{\textbf{Top 10}} \\ \hline
\textbf{MS N-CalTech (Face)}     & 95                            & 100                           & \multicolumn{1}{c}{100}             \\ \hline
\end{tabular}%
}
\end{table}

\subsection{Object Detection and Segmentation in a Multi-Class Environment }

\begin{figure}[hbt!]
\begin{center}
\includegraphics[width=0.98\linewidth]{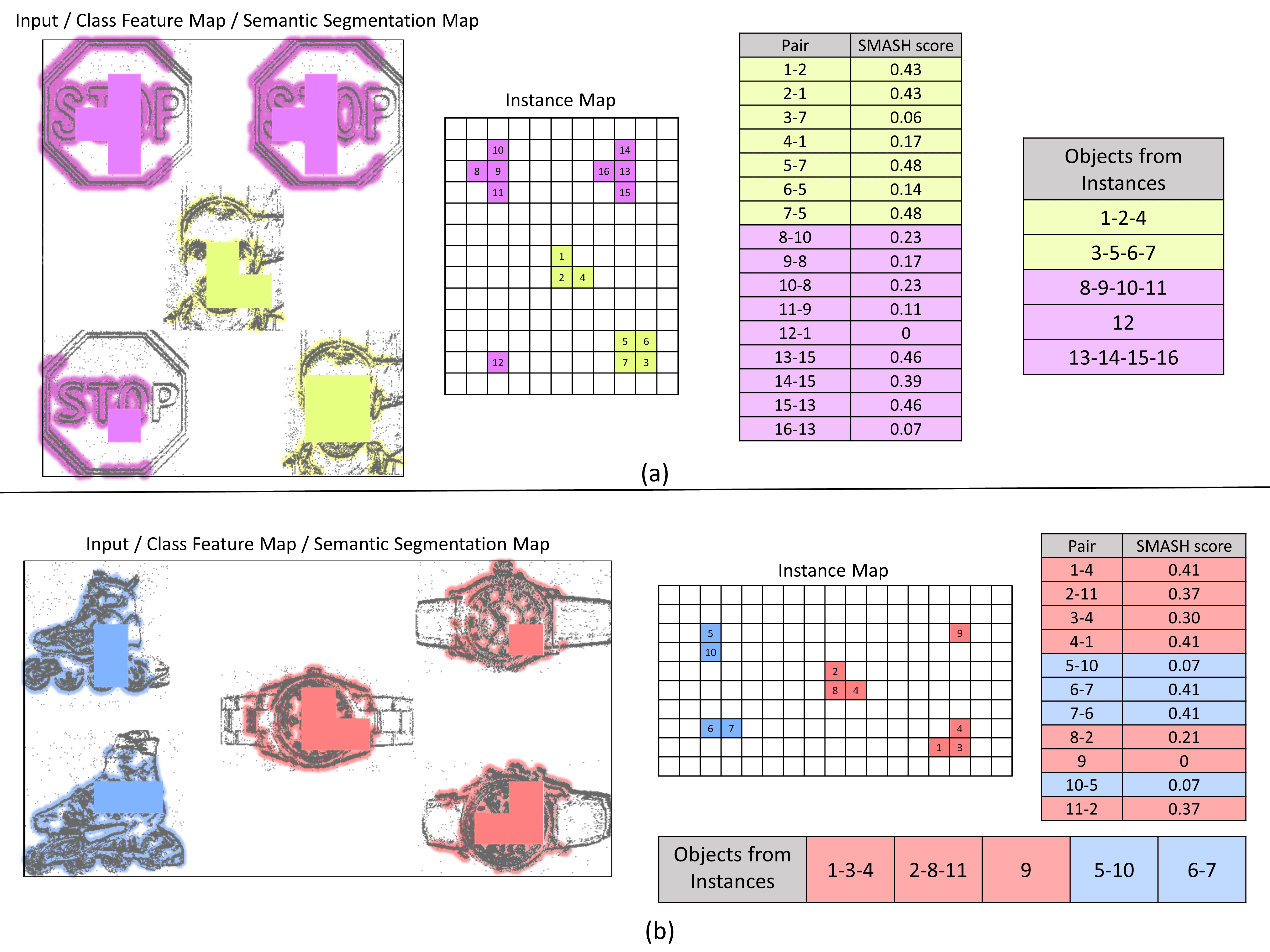}
\end{center}
\caption{Illustrating the process of going from semantic segmentation to instance wise object construction, displaying the instance mapping and pairing with SMASH scores that help to construct the class instance objects for two examples (a) and (b). Respectively showing the process for the classes Face and Stop Sign, and Watch and In-line Skate.} 
\label{fig:5classinstance}
\end{figure}

To further evaluate the SMASH method, the multi-class semantic segmentation networks were tested. This allowed testing of the object detection and instance segmentation from the 5 and 10 class semantic segmentation environment. Together with the increased number of classes present in each segment, there was also a reduction in the overall number of features available within the second convolution layer, Conv 2. This presents a scenario where the features are more generalised and therefore present less opportunities to learn features that help to diversify. Despite this, the SMASH process was still able to deliver high accuracy in determining how many object instances of a class there were, with 95.17\% within the 5 class testing and 91.67\% for the 10 class test, collated in Table \ref{tab:my-table}. 
To help visualise the results from this experiment, two examples of successful instance segmentations outputs from the 10 class test are shown in \figurename{} \ref{fig:5classinstance}, which shows for both (a) and (b), the input overlaid with the classification layer feature map and the semantic segmentation output respectively. \figurename{} \ref{fig:5classinstance} also indicates how the class instances are assigned in the `Instance Map', where the numerical assignment is displayed to help illustrate the SMASH score also seen within the image. The Pairs are shown in numerical order with associated SMASH score, and how that is grouped into objects.


The classes other than Face presented a different problem to the Face class, as most of the other classes (excluding Stop Sign) had a higher feature count in the second convolution layer Conv 2. This presented a problem in which the timing of the occurrences of the features, was more important than its occurrence. Which during the Face only testing, was not an issue due to the sparsity at which those features appeared. This testing highlights the importance of the temporal component of the similarity matching, as simple feature-wise similarity matching alone is unable to determine class instances from one another. This extra temporal dimension can be seen a being able to leverage the saliency (timing) of features within individual class instances.

\FloatBarrier

\subsection{Object Occlusion}
\begin{figure}[h]
\begin{center}
\includegraphics[width=0.98\linewidth]{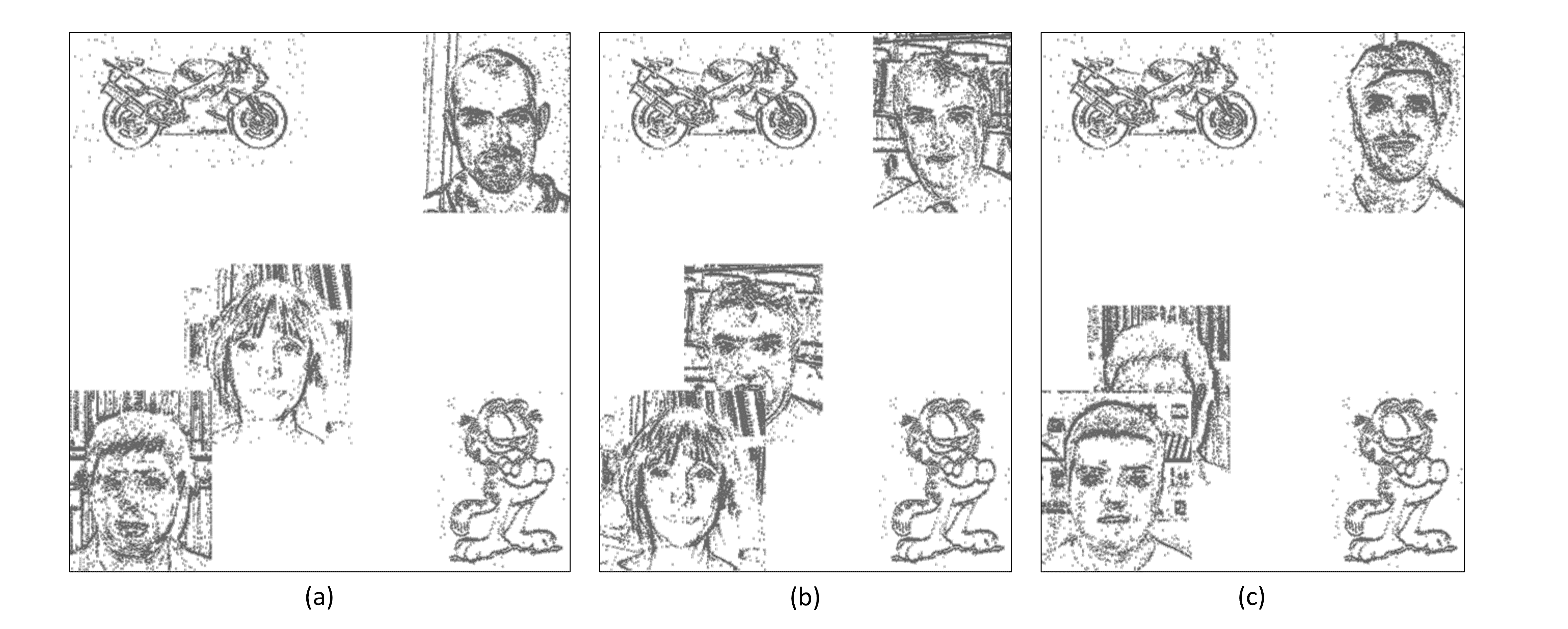}
\end{center}
\caption{Input occlusion example illustrating: (a) 5\% occlusion of center image, (b) 25\% occlusion of center image, and (c) 50\% occlusion of center image} 
\label{fig:inputocclusion}
\end{figure}

To help test the HULK-SMASHs ability to be able to perform sequence to sequence tracking of objects, further evaluation of the object detection capability is tested within three more challenging scenarios, where the class instance will now overlap. Taking the same multi-stream approach as the previous test, however, now the central stream of the image will be positioned such that occlusion of 5\%, 25\% and 50\% will occur. Partial occlusions test the feature extraction and similarity matching performance when there is a bounding box overlap, meaning the SMASH score is now the active measure of the combined similarity and locality. The three occlusion states are illustrated in \figurename{} \ref{fig:inputocclusion}, where (a), (b) and (c) depicted the occlusion covering for 5 \%, 25\% and 50\% respectively as the stream of data in the centre is repositioned to create the occlusion.

%

\begin{figure}[hbt]
\begin{center}
\includegraphics[width=0.75\linewidth]{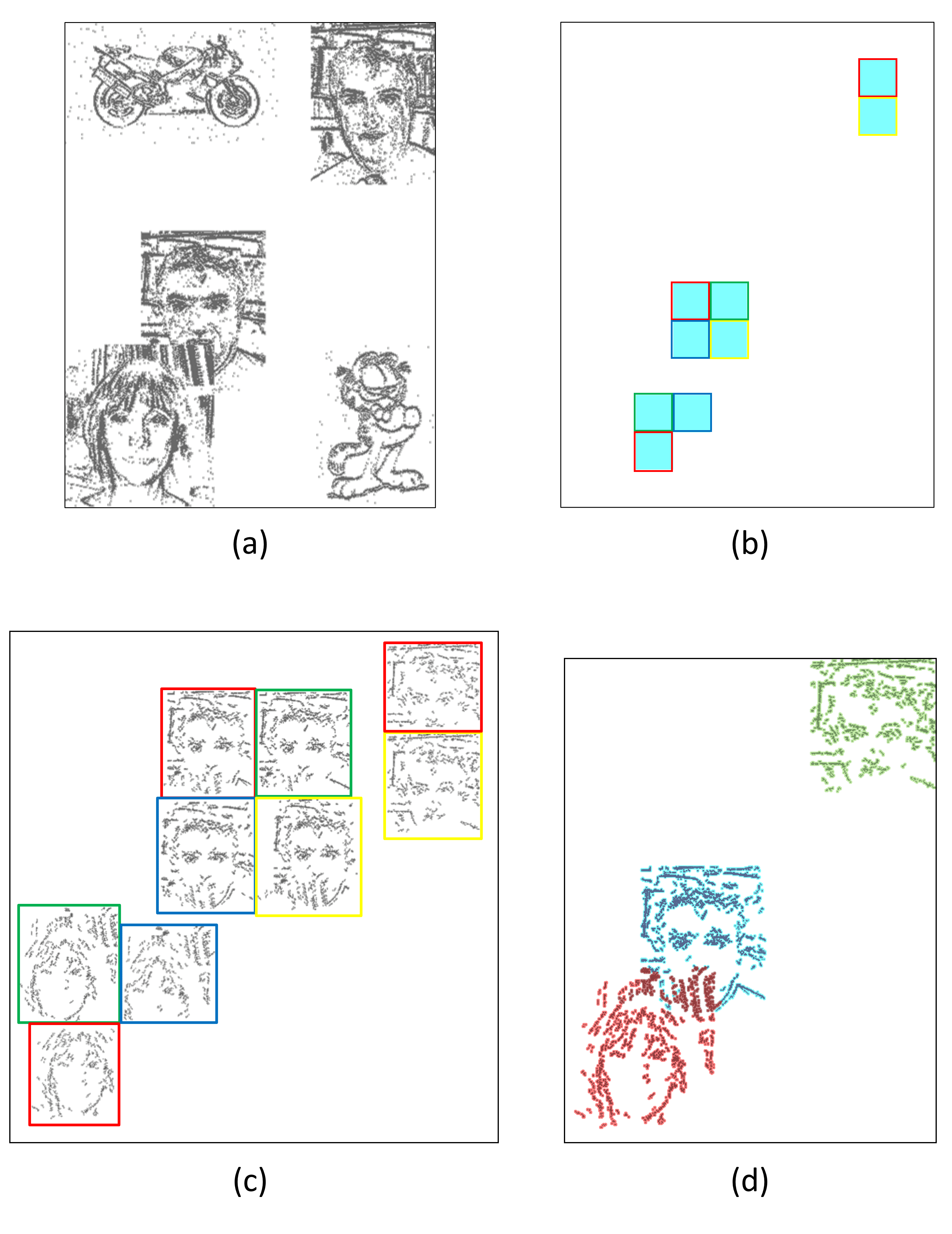}
\end{center}
\caption{Input 25\% occlusion example illustrating: (a) The input sequence, (b) The latent classification space, (c) The class instance breakdown and (d) The instance segmentation output} 
\label{fig:25occlusion}
\end{figure}

\begin{figure}[hbt]
\begin{center}
\includegraphics[width=0.60\linewidth]{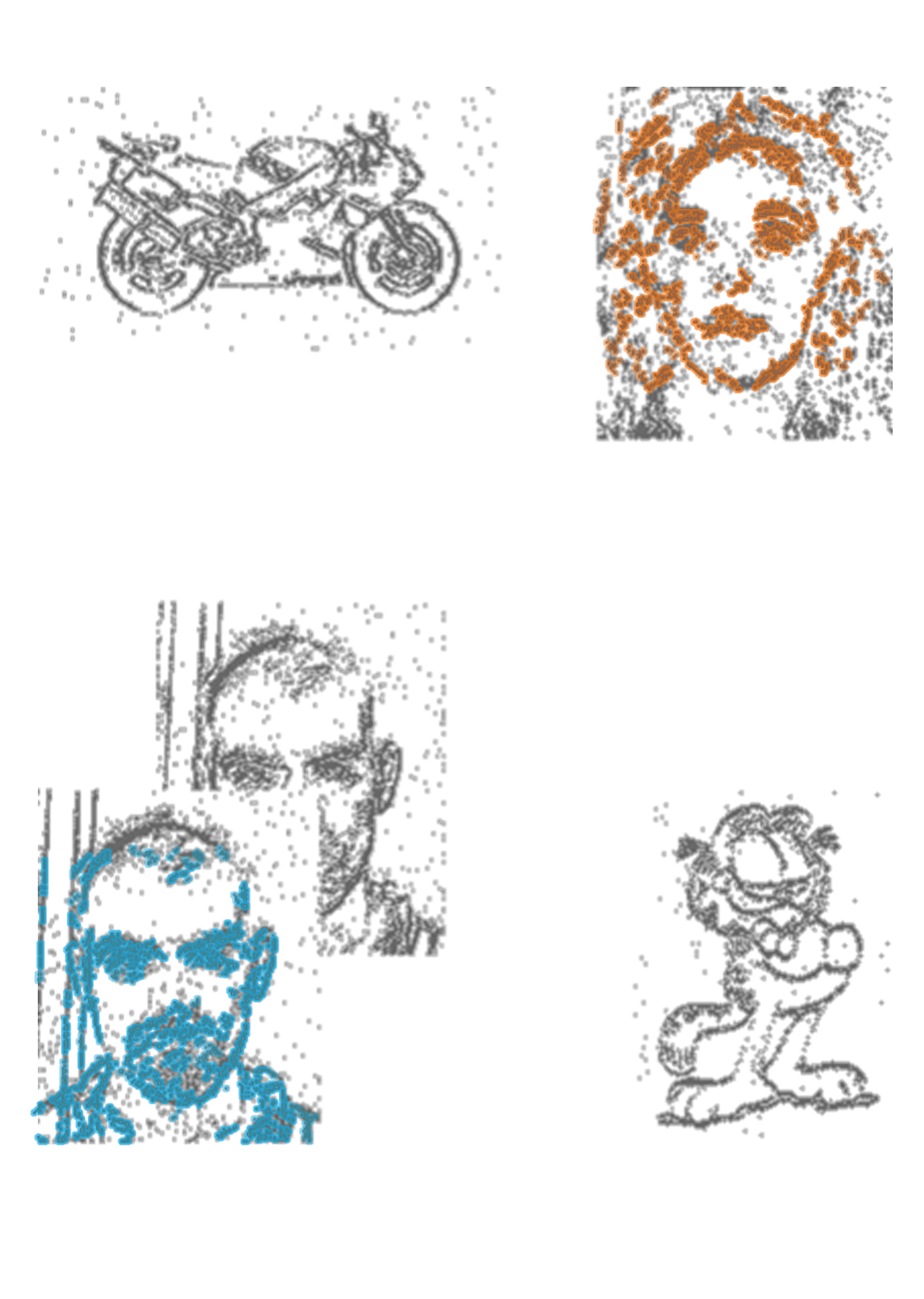}
\end{center}
\caption{Segmentation mapped over the input sequence, with a failure to identify the central image, thus failure to include in instance segmentation, however successful segmentation of the other two faces} 
\label{fig:failocclusion}
\end{figure}

\noindent With the 5\% occlusion, the performance was unchanged at 97.33\% also shown in Table \ref{tab:my-table}, which was to be expected considering the example shown in \figurename{} \ref{fig:inputocclusion} (a) shows no overlapping of the facial features, with all images in the dataset having the faces centred in the image. While the 25\% occlusion reduced the performance down to 62\% and with 50\% occlusion getting 28\% also both in Table \ref{tab:my-table} when testing over each input buffer. 
A more detailed example from the 25\% occlusion testing is shown in \figurename{} \ref{fig:25occlusion}. The figure highlights the process from the input in (a), the class latent space neurons that were active in (b), the HULK end process of the active class pixels represented back in the pixel space with (c), and finally, the instance segmentation output (d), with three successful instances of each face detected.
The only concern with this approach is the overlapped section between the bottom two images has a region that is assigned to two objects. This results in an area of uncertainty and even with a process where the two objects could compete for the pixels, the misclassification appears to exist in the feature detection domain, not the HULK-SMASH process. However, it must be noted that the failures within this testing phase are exclusively from the failure to find enough features to give a positive classification, not the incorrect assignment of instances. This element is highlighted in \figurename{} \ref{fig:failocclusion}, where the classification process has failed to result in classification for the centre image; however, the bottom left and top right faces are correctly classified and segmented without overlap onto the unclassified region.

\FloatBarrier
\subsection{Object Occlusion Recovery}
\begin{figure}[hbt]
\begin{center}
\includegraphics[width=0.98\linewidth]{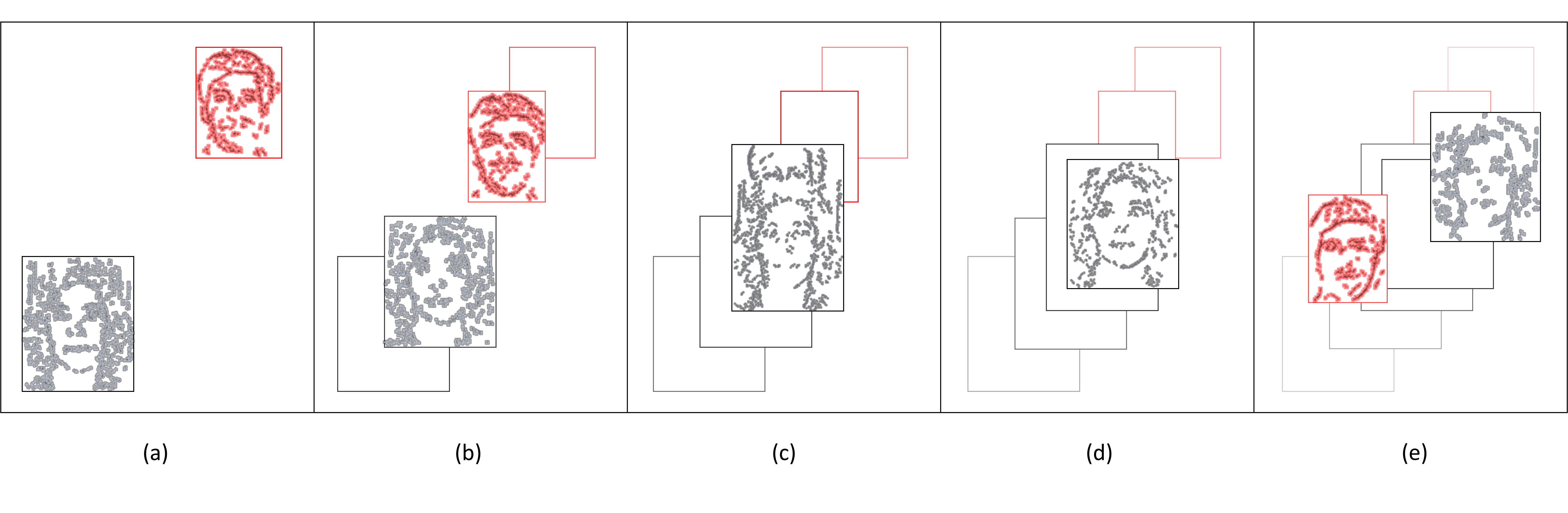}
\end{center}
\caption{Segmentation mapped over the input sequence, with a failure to identify the central image, thus failure to include in instance segmentation, however successful segmentation of the other two faces} 
\label{fig:tracking}
\end{figure}

As shown in the previous section, the object detection and instance segmentation method can identify features belonging to a particular instance of that class on an input by input basis, known as intra-sequence detection. The experimental set up for this section allows testing of object detection across inputs multiple sequences known as inter-sequence detection.
This test then allows sequence to sequence tracking, then partial and full occlusion and subsequent recovery. The inputs are the full 300 ms sequence of the N-Caltech dataset \cite{Orchard2015converting} broken into the usual 10 ms steps as seen in previous sections. 
This places two inputs at the top right and bottom left, then has then move diagonally across towards each other for each buffered input. 
The results of the test from a few selected instances are shown in \figurename{} \ref{fig:tracking}. \figurename{} \ref{fig:tracking} (a) depicts the start position of the two different sequences and their respective instance segmentation and bounding boxes (red and black). \figurename{} \ref{fig:tracking} (b-e) show the transition of the respective segmentations diagonally across, with (c) and (d) showing the occlusion due to overlap where the black object completely occludes the red object. \figurename{} \ref{fig:tracking} (e) then highlights the detection picking up the red object again, from finding a high similarity with the red object that existed prior to the occlusion.

\begin{figure}[hbt]
\begin{center}
\includegraphics[width=0.8\linewidth]{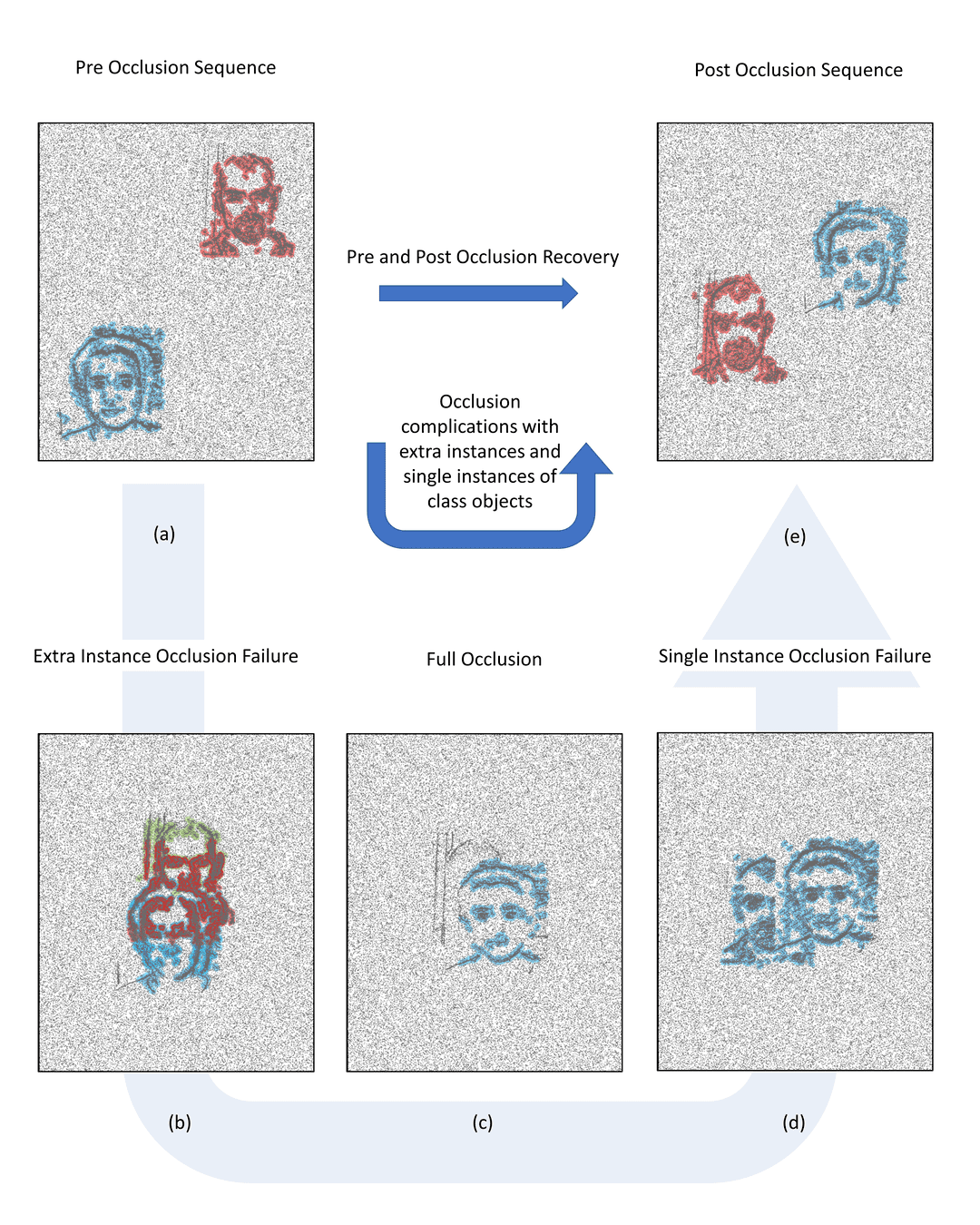}
\end{center}
\caption{Occlusion recovery with noisy input. (a) Pre Occlusion state, (b) Occlusion causing extra objects to appear, (c) Full occlusion with only one object present, (d) Partial occlusion grouping of objects and (e) Post occlusion state} 
\label{fig:occlusionnoise}
\end{figure}

The experimental results show the inter-sequence similarity matching is able to recover the occluded input sequence, such that across all the test sequences the similarity check managed to recover the occluded objects 100\% of the time, as seen in Table \ref{tab:my-table}. This feat is more impressive due to the occluded object being a dynamic sequence itself, meaning the post occlusion object is not just a replica of pre occlusion one. 
This occlusion recovery was also tested again with a noisy input, that is added noise to both the input sequence and to the general background. This noisy sequence is shown in \figurename{} \ref{fig:occlusionnoise}, where \figurename{} \ref{fig:occlusionnoise} (a) and (e) shown the pre and post occlusion states, presenting that the recovery works correctly, while \figurename{} \ref{fig:occlusionnoise} (b), (c) and (d) show some of the issues the occlusion can cause. \figurename{} \ref{fig:occlusionnoise} (b) highlights the failure of the object detection in finding extra objects that do not exist. \figurename{} \ref{fig:occlusionnoise} (c) correctly illustrates the full occlusion state where only the female face object is present. \figurename{} \ref{fig:occlusionnoise} (d) displays the point at which partial occlusion is still causing only one object to be present, due to the similarity of class instances being close across the 9 instances in this case. Some of the 9 instances include multiple features of both objects, similar to what is shown in \figurename{} \ref{fig:occlusionnoise} (b). However, in this case, there was at least one instance that matched another within both objects. The object in \figurename{} \ref{fig:occlusionnoise} (d) is also blue as it scored the highest similarity with that object on the inter-sequence process. 
Despite some of the mentioned inaccuracies associated with the added noise, the system was able to re-identify occluded objects 100\% of the time, as shown in \figurename{} \ref{fig:occlusionnoise} with this result recorded with the others in Table \ref{tab:my-table}. 

\FloatBarrier
\section{Intuition and Understanding through Visualisation}
\label{ch:TSM:sec:vis}
\begin{figure}[hbt]
\begin{center}
\includegraphics[width=0.8\linewidth]{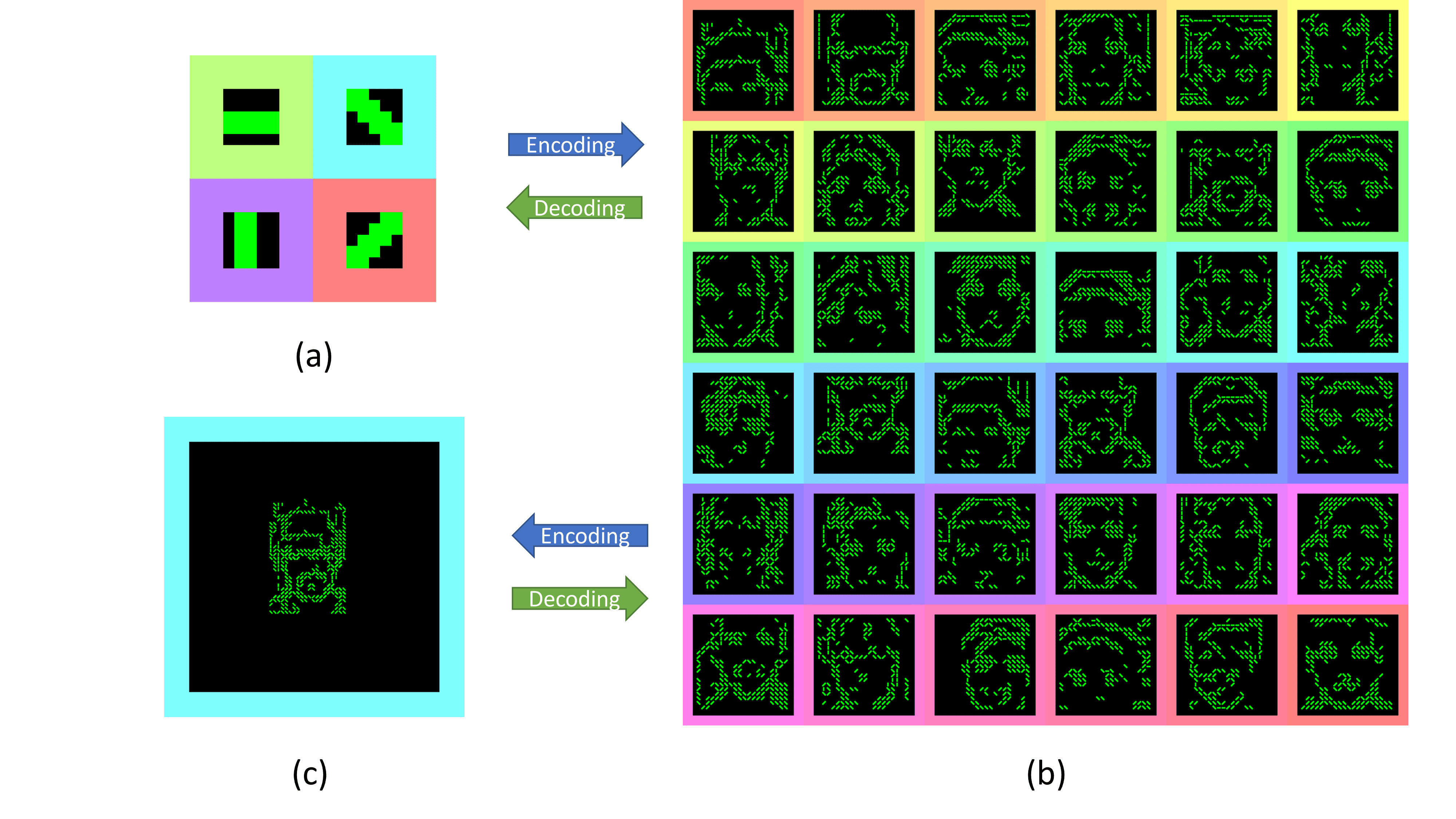}
\end{center}
\caption{The set of features from the trained network, (a) Conv 1, (b) Conv 2 and (c) Conv3} 
\label{fig:featureset}
\end{figure}

The principle behind how object detection and instance segmentation are resolved can be better understood through visualisation. Through visualisations of the weights as they are mapped back into the pixel space through the subsequent layers are shown in \figurename{} \ref{fig:featureset}. \figurename{} \ref{fig:featureset} (a) showing the features of Conv 1, which are the unlearned pre-determined edge detection features, (b) showing the features of layer Conv 2 (Trans-Conv 2) and (c) highlighting the classification layer feature which depicts a Face. \figurename{} \ref{fig:featureset} (b) contains the bulk of the information used for the similarity matches process, thus contains the key to interpreting how the network can differentiate between different people in the dataset.

\begin{figure}[hbt]
\begin{center}
\includegraphics[width=0.8\linewidth]{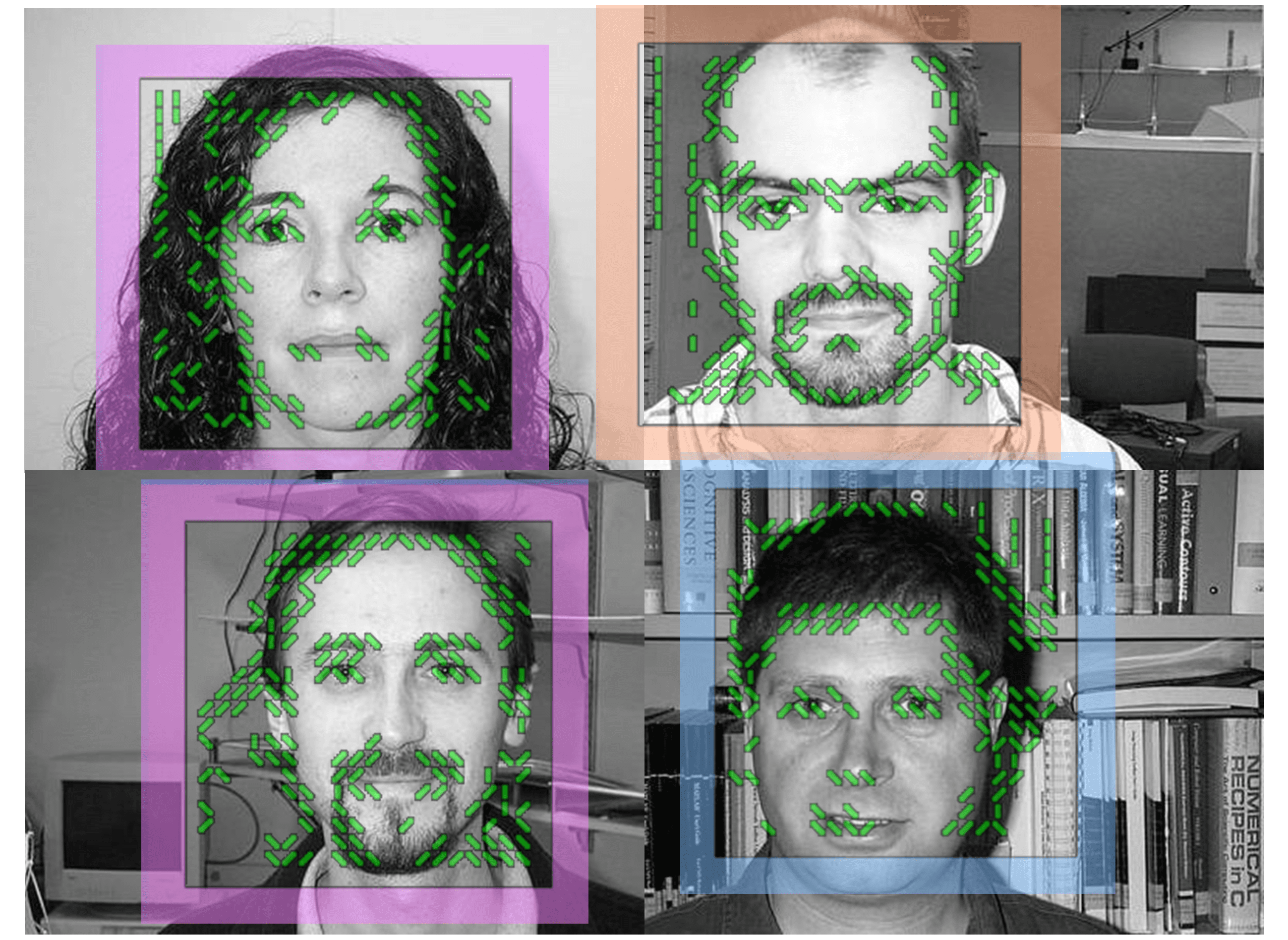}
\end{center}
\caption{Overlaid Features from Conv 2 onto Images from original Caltech Dataset, displaying how the learned features represent different faces} 
\label{fig:featureoverlay}
\end{figure}

Mapping the features of the network from Conv 2 layer onto some original images from the Caltech dataset presents an insight into how this layer is differentiating between the different people in the dataset, as each person has unique and distinctive enough features to be learned through repeat occurrence. This ability to learn sub-classification feature clustering stems from the ability to classify up to 10 different objects due to the variance between the classes (inter-class) being higher than the variance of one particular class (intra-class). However, in this case, it is now the intra sub-class variance that allows the unique featural-temporal identifiers of each face to be captured.
This is highlighted in \figurename{} \ref{fig:featureoverlay} where 4 distinctive features that appear to represent a particular person from the dataset are overlaid onto the non-spiking version of the original image (for easier visualisation) it matches closest. In which the features of \figurename{} \ref{fig:featureoverlay} appear to pick out the typical facial features of eyes nose mouth and hair.

\begin{figure}[hbt]
\begin{center}
\includegraphics[width=0.8\linewidth]{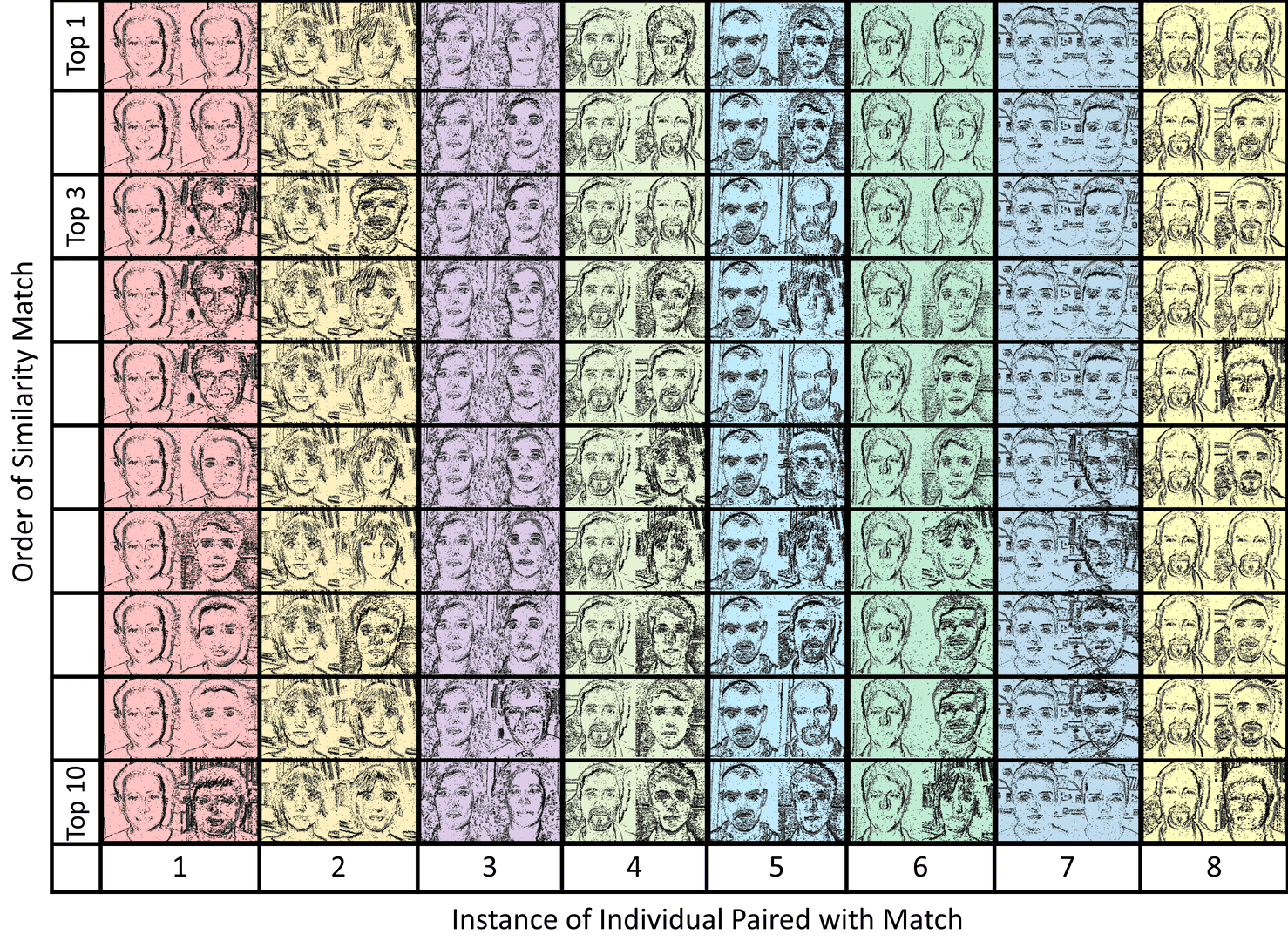}
\end{center}
\caption{Columns of top 10 matches, left on each column is the test input repeated 10 times with each of the 10 matches to the right, ranked in descending order. Each colour represents a new column for easier visual separation} 
\label{fig:people}
\end{figure}

To test just how well the features are allowing the individuals within the dataset to be recognised, an experiment matching each sequence with its top 1, top 3 and top 10 matches was carried out. This permits an insight into whether or not the HULK-SMASH process was able to differentiate between the people within the dataset, purely on re-occurrence of similar salient features, or in other words distinctive facial features in a featural-temporal manner. The test processes each buffered input, within each sequence of the testing set, and compares it with the full test set, recording the top, top 3 and top 10 most similar spatial-temporal patterns to its own. The results show that over 95\% of the test inputs for an individual match the highest with themselves. While 100\% match with themselves if extended into the top 3 and top 10, displayed in table \ref{tab:my-table}. This result is illustrated in \figurename{} \ref{fig:people}, where the top 10 similarity matches for a selection of inputs are shown in a coloured column-wise manner, with the left of the column showing the input sequence and the right showing the top 10 matches in descending order.

\figurename{} \ref{fig:people} helps to build interpretability to the similarity matching process, allowing an understanding as to how and why the accuracy of the classification and object detection is as such. The sparse feature sets of a winner takes all approach to STDP allowing easier visualisation. This in turn helped to construct a method of matching these spatial-temporal features which are a result of the neuromorphic input to the spiking network both maintaining the important temporal causality of salient features. 
\FloatBarrier

\section{Conclusion}
\label{ch:TSM:sec:conclusion}
In this paper a new method for spiking object detection and instance segmentation, HULK-SMASH was presented. The system utilised STDP learned features that previously gave a semantic segmentation output. The classification layer was subsequently processed by unravelling each individual latent spiking neuron in the layer back to the pixel space, seen as HULK. This allows each instance of the classification layer to be treated as a class instance. It is then through comparing the class instances that objects can be identified through the SMASH score looking at the similarity and locality of instances. Permitting object detection and instance segmentation to be carried out on each buffered input. Then through similarity matching of the objects in sequences, sequence to sequence object detection and segmentation was achieved. This allowed object occlusion and reappearance to be realised, with the ability to deal with subtle changes to the object pre and post occlusion. The same mechanisms for SMASH could be used for continual online learning in which new objects are created so long as they trigger a classification neuron. Visualisation of experimental results gave a better understanding and intuition as to where the network was working well and why some failures occur. The HULK-SMASH processes were shown to be robust to multiple network instances with single and multi-class experiments with successful results while aiding in neural network interpretability. 

\FloatBarrier







\bibliographystyle{ieeetr}

\end{document}